\pdfoutput=1

\documentclass[11pt]{article}

\usepackage[]{EACL2023}

\usepackage{times}
\usepackage{latexsym}

\usepackage[T1]{fontenc}

\usepackage[utf8]{inputenc}

\usepackage{microtype}

\usepackage{inconsolata}

\usepackage{amsmath} 
\usepackage{amssymb}

\usepackage{algorithm}
\usepackage{xcolor}
\usepackage{algpseudocode}

\usepackage{graphicx}
\usepackage{multirow}

\usepackage{enumitem}

\title{ComSearch: Equation Searching with Combinatorial Strategy for Solving Math Word Problems with Weak Supervision}

\author{Qianying Liu$^1$, Wenyu Guan$^2$, Jianhao Shen$^3$, Fei Cheng$^1$ and Sadao Kurohashi$^1$\\
$^1$ Graduate School of Informatics, Kyoto University \\
$^2$ Xiaomi AI Lab\\
$^3$ Peking University\\
  {\tt ying@nlp.ist.i.kyoto-u.ac.jp; 
  gwy1995@live.com; {jhshen}@pku.edu.cn;} \\
  \tt \{feicheng,kuro\}@nlp.ist.i.kyoto-u.ac.jp;
  }

\begin{document}

\setlength{\abovedisplayskip}{0pt}
\setlength{\belowdisplayskip}{0pt} 
\setlength{\abovedisplayshortskip}{0pt}
\setlength{\belowdisplayshortskip}{0pt}
\setlength{\textfloatsep}{3pt}

\maketitle
\begin{abstract}
Previous studies have introduced a weakly-supervised paradigm for solving math word problems requiring only the answer value annotation. While these methods search for correct value equation candidates as pseudo labels, they search among a narrow sub-space of the enormous equation space. To address this problem, we propose a novel search algorithm with combinatorial strategy \textbf{ComSearch}, which can compress the search space by excluding mathematically equivalent equations. The compression allows the searching algorithm to enumerate all possible equations and obtain high-quality data. 
We investigate the noise in the pseudo labels that hold wrong mathematical logic, which we refer to as the \textit{false-matching} problem, and propose a ranking model to denoise the pseudo labels. 
Our approach holds a flexible framework to utilize two existing supervised math word problem solvers to train pseudo labels, and both achieve state-of-the-art performance in the weak supervision task. \footnote{Our code and data is available at \url{https://github.com/yiyunya/ComSearch}}

\end{abstract}

\section{Introduction}

Solving math word problems (MWPs) is the task of extracting a mathematical solution from problems written in natural language. 
Based on a sequence-to-sequence (seq2seq) framework that takes in the text descriptions of the MWPs and predicts the answer equation~\cite{wang-etal-2017-deep}, task-specialized encoder and decoder architectures~\cite{wang2018mathdqn, wang2019template, DBLP:conf/ijcai/XieS19, liu-etal-2019-tree, guan-etal-2019-improved, zhang2020graph, ijcai2020-555, shen-jin-2020-solving}, data augmentation and normalization~\cite{wang2018translating,liu2020reverse,shen-etal-2022-seeking}, and pretrained models~\cite{tan2021investigating,liang2021mwpbert,shen2021generate,shen-etal-2022-textual} 
have been conducted on \textit{full supervision} setting of the task. 
These settings require equation expression annotation, which is expensive and time-consuming.
  \begin{figure}[t]
  \centering
  \includegraphics[width=0.50\textwidth]{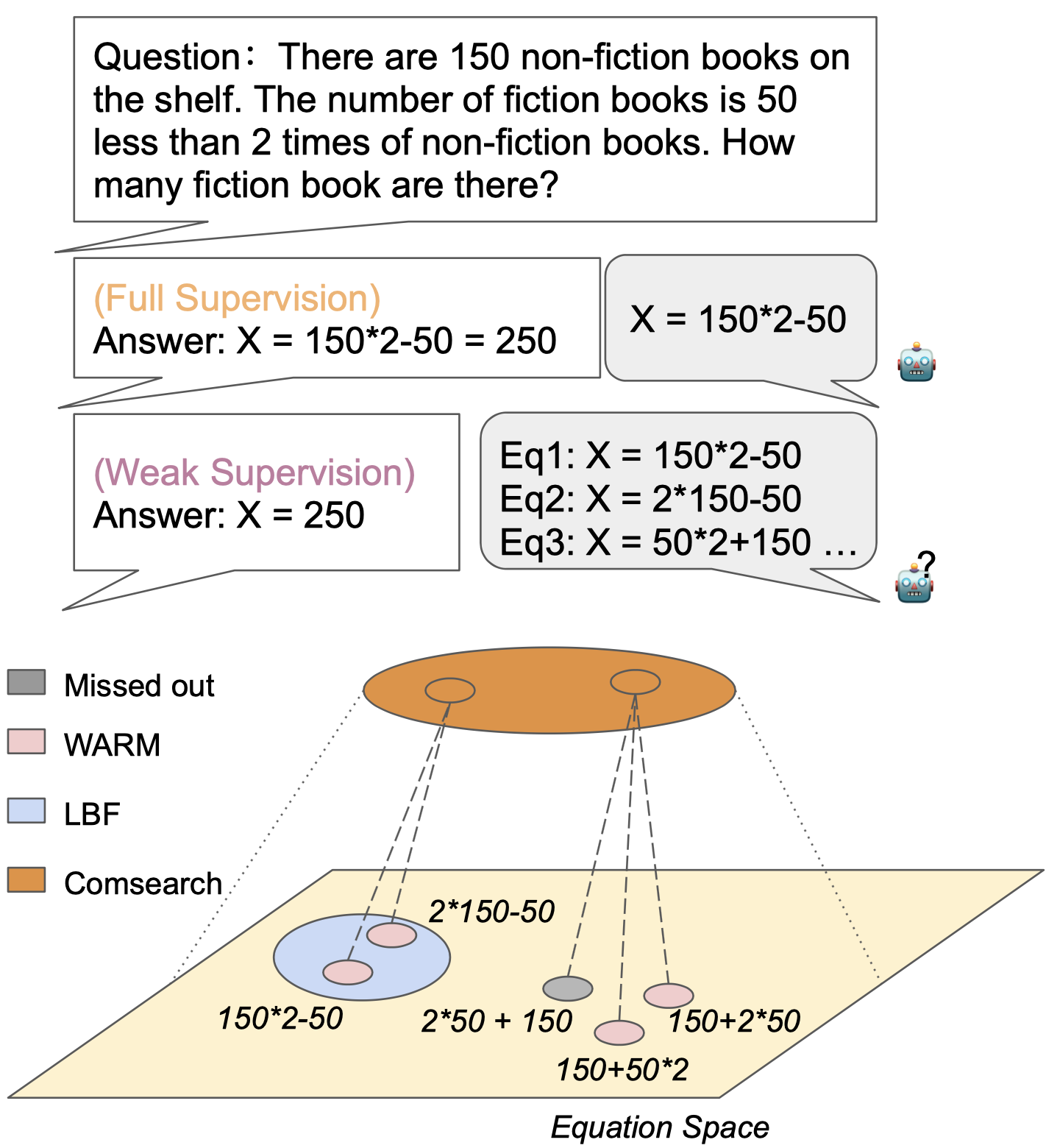}
  \caption{Example of MWP solving system under full supervision and weak supervision.}
  \label{fig:ex}
  \end{figure}

Recently \citet{Hong_Li_Ciao_Huang_Zhu_2021} (LBF) and \citet{chatterjee2021weakly} (WARM) addressed this problem and proposed the \textit{weak supervision} setting, where only the answer value annotation is given for supervision. 
Such a setting forms pseudo question-candidate equation pairs, which hold the correct answer value for training with the complexity of $O(n^{2n})$ for $n$ variables enormous possible equation space. Computational efficiently extracting such pairs becomes the major challenge since it is computationally impossible to traverse all possible equations, especially when the example has more variables (e.g., 88,473,600 for 6 variables). As we show in Figure \ref{fig:ex}, previous studies sample a limited set of equations via random walk \cite{Hong_Li_Ciao_Huang_Zhu_2021} or beam searching \cite{chatterjee2021weakly}.
However, the algorithms can only cover a limited part of the data, which we refer to as \textit{recall}.
As shown in Table \ref{tab:long_short}, LBF \cite{Hong_Li_Ciao_Huang_Zhu_2021} only covers 30\% of the examples of more than 4 variables. Moreover, the random walk algorithm lacks robustness and leads to a high performance variance.

\begin{table}[t]
\centering
\begin{tabular}{l|rr}
\hline
\textbf{Model}& $\leq 3$ & $\geq 4$ \\
\hline
LBF & 88.1\% & 30.9\% \\
ComSearch & 94.4\% & 94.5\%\\
\hline

\end{tabular}
\caption{Searching result recall on problems of different variable sizes.}
\label{tab:long_short}
\end{table}

We observe that although the equation search space is ample, many equations are mathematically equivalent 
under the commutative law, associative law, or other equivalent forms. Hence, searching for these equivalent equations is redundant, especially for difficult examples with a larger number of variables. For example, $a+b+c+d*e$ has 48 equivalent forms that hold the same mathematical meaning considering only the commutative law. Eliminating such redundancy in the searching space could reduce computational complexity. 
In this paper, we propose 
a combinatorial-strategy-based searching method \textbf{ComSearch} that enumerates \textit{non-equivalent equations} without repeating, which can robustly extract candidate equations for a wide range of unlabeled data and build a high recall pseudo data with equation annotation even for difficult examples.
To this end, the main idea of Comsearch is to use depth-first search (DFS) to enumerate only one representative equation for each set of equivalent equations and then check whether the equation holds the correct answer value.
Comsearch effectively compresses the searching space, e.g., up to 111 times for 6 variables compared to bruce-force searching. 
As shown in Table \ref{tab:long_short}, ComSearch can achieve a relatively high recall for different variable sizes.
Our method could be proven to have lower approximate complexity.

While Comsearch only searches among \textit{non-equivalent equations}, we observe that many examples still have multiple candidate equations through which we can get the final answer.
As shown in Figure 1, Equation 1 (Eq1: X=150*2-50) and Equation 3 (Eq3: X=50*2+150) can get the same value, but Equation 3 holds a false mathematical reasoning logic, and using Equation 3 as the pseudo label would bring in noise. We address this data noise as the \textit{false-matching} problem, which has been ignored in previous studies,
since their methods do not consider whether the multiple candidate equations of one example are caused by equivalent equation forms or false matching.
To address this problem,
we investigate how the false-matching problem drags down the system's performance and propose two ranking models to alleviate this problem. 
For examples with multiple candidate equations from ComSearch, the ranking module first collects a set of candidate equations, then assign a score by a draft model trained on pseudo data with only a single candidate equation to each candidate to choose the best pseudo label. In addition to candidates from the searching result of ComSearch, we observe that beam search results of the draft model can also serve as a high-precision candidate equation. We investigate these two settings for candidate equation sets.  

We conduct experiments on two strong MWP solvers, achieving state-of-the-art (SOTA) results under the weakly supervised setting, especially for examples with many variables.
The results also demonstrate the effectiveness and generalization ability of our method.

\begin{figure*}[t]
  \centering
  \includegraphics[width=0.9\textwidth]{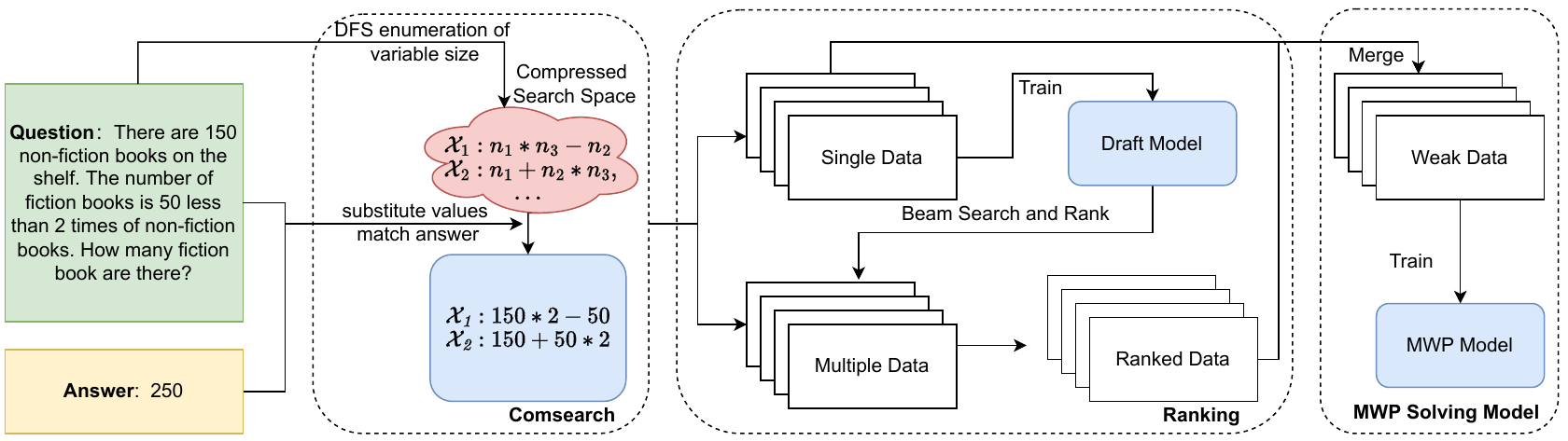}
  \caption{The model overview. }
  \label{fig:model}
  \end{figure*}

In summary, our contribution is three-fold:
\begin{itemize}[noitemsep]
    \item We propose ComSearch, a searching algorithm that enumerates non-equivalent equations without repeat to search candidate equations effectively.
    \item We are the first to investigate the \textit{false-matching} problem that brings noise to the pseudo training data. We propose a ranking module to reduce the noise and give a detailed oracle analysis of the problem.
    \item We perform experiments on two MWP solvers with our ranking module and achieve SOTA performance under weak supervision.
    
\end{itemize}

\section{Methodology}

We show the pipeline of our method in Figure \ref{fig:model}. Our method consists of three modules: the Search with combinatorial strategy (\textbf{ComSearch}) module that searches for candidate equations; the MWP model that is trained to predict equations given the natural language text and pseudo labels; 
the Ranking module that uses an explorer model to find candidate equations and select the best candidate equation with a scoring model.

\subsection{ComSearch}

Directly searching for non-equivalent equation expressions is difficult because the searching method needs to consider Commutative law, Associative law, and other equivalent forms. We show how equivalent equations could be merged into a representative form $\mathcal{X}$, and the enumeration of $\mathcal{X}$ can transverse all non-equivalent equations for four arithmetic operations.

We define the set of \textit{non-equivalent equations} using four arithmetic operations as $S_n$. We first split the equations to two categories, either $S^{\pm}$ where the outermost operators are $\pm$, such as $n_1/n_2+n_3-n_4$ and $n_1/n_2-(n_3-n_4)$, or $S^{\divideontimes}$ where the outermost operators are $\divideontimes$, such as $(n_2+n_1)*(n_3-n_4/n_5)$. We call the former a \textit{general addition equation} and the latter a \textit{general multiplication equation}.

\begin{align}
\label{eq:gaset}
    S^{\pm}_m=\{(n_1\divideontimes(..)) \pm (n_{i} \divideontimes (..)) \pm.. n_m\}\\
    \label{eq:gmset}
    S^{\divideontimes}_m=\{(n_1 \pm (..)) \divideontimes (n_{i} \pm (..)) \divideontimes.. n_m\}
\end{align}

These two sets are symmetrical, so we only need to consider one set. Consider elements in $S^{\pm}_m$, we can rewrite the equation to the representative form $\mathcal{X}$:
\begin{align*}
    \mathcal{X} =& ((n_i\divideontimes(..)) + (n_{j} \divideontimes (..)) + ..) \\&- ((n_{k}\divideontimes(..)) + (n_{l}\divideontimes(..)) + ..)  \end{align*}
    
For example, $n_1/n_2-n_3+n_4$ and $n_1/n_2-(n_3-n_4)$ are equivalent, that they are both rewritten as $(n_1/n_2+n_4)-n_3$. $(n_2+n_1)*(n_3-n_4/n_5)$ could be rewritten as $(n_1+n_2)*(n_3-n_4/n_5)$. 
Trivially, any two equations that are represented by the same $\mathcal{X}$ are equivalent.
We give proof of the number of inequivalent expressions involving $n$ operands in Appendix Section \ref{sec:proof}, which shows that any two equivalent equations are written as the same $\mathcal{X}$.
Thus the enumeration of $\mathcal{X}$ is equivalent to the enumeration of non-equivalent equations. 
The enumeration problem of these equations is an expansion of solving Schroeder's fourth problem~\cite{sch1870}, which calculates the number of labeled series-reduced rooted trees with $m$ leaves. We give the details of the DFS in the Appendix Section \ref{sec:alg}.

Given the compressed search space, we substitute the values for variables in the equation templates and use the equations of which value matches the answer number as candidate equations. If no equations could be extracted by using all numbers, we continue to consider: (1) omitting one number, (2) adding constant number 1 and $\pi$, and (3) using one number twice. If the algorithm extracts candidates at any stage, the further stages are not considered since it would introduce repeating equations, e.g., $1*(a+b)$ is a duplication of $a+b$.

\subsection{MWP Solving Models}


\paragraph{Goal-driven Tree-structured Solver}
We follow \citet{Hong_Li_Ciao_Huang_Zhu_2021} and \citet{chatterjee2021weakly} and use Goal-Driven Tree-Structured MWP Solver (GTS) 
\cite{DBLP:conf/ijcai/XieS19} as the MWP model.
GTS is a seq2seq model with the attention mechanism that uses a bidirectional long short term memory network (BiLSTM) as the encoder and LSTM as the decoder. GTS also uses a recursive neural network to encode subtrees based on its children nodes' representations with the gate mechanism. With the subtree representations, this model can well use the information of the generated tokens to predict a new token. 

\paragraph{Graph-to-Tree Solver}

Following \citet{chatterjee2021weakly}, we conduct experiments on Graph-to-Tree Solver (G2T)~\cite{zhang2020graph} . G2T is a direct extension of GTS, which consists of a graph-based encoder capturing the relationships and order information among the quantities.

\subsection{Ranking}
\label{sec:rank}

While ComSearch enumerates equations that are non-equivalent without repeat, some variable sets can coincidentally form multiple equations with the same correct value, as shown in Figure \ref{fig:model}. The equations $150*2-50$ and $150+50*2$ are non-equivalent. However, their values are equal, while only $150*2-50$ is the correct solution. We refer to this problem as \textit{false-matching}, an important issue that previous studies have overlooked. While previous studies also collect multiple candidate equations for one example, they cannot differ whether the issue is caused by equivalent forms of the equations or \textit{false-matching}, and they do not perform any processing on these \textit{false-matching} examples, which brings in noise to the pseudo data.

To process these data that have multiple candidate equations, we propose two ranking methods to choose the best candidate equation for each example.  
The module first collects a set of candidate equation that holds the correct annotated answer value and then score the candidates to choose the pseudo label for the sample.

Before ranking, we train a draft model $S$ on the single-candidate pseudo data because the single-candidate data is relatively reliable with fewer false-matching examples. In the first ranking method \textit{Basic Ranker}, for a data example $x$, 
we rank among the multiple search results of Comsearch $\{y^{eq}\}^{search}$. Then we use the draft model $S$ to calculate the conditional probability of $y^{eq}$ at each time step $t$. The score of the length $k$ equation $s^{eq}$ is defined as:
\begin{equation}
    s^{eq} = \sum_{t=0}^k log(S(x, y_t^{eq}))
\end{equation}

We use the candidate equation that has the highest score as the pseudo label of this example.

Empirically, we observe that performing beam search on the draft model $S$ could also generate high-precision candidate equations. 
Thus in the second method \textit{Beam Ranker}, we further explore more candidate equations with beam search. We add beam search predictions of $S$ that hold the correct value $\{y_{eq}\}^{beam}$ to the candidate equation set along with Comsearch results $\{y^{eq}\}^{search}$. The score function is defined the same as the basic ranker.

\begin{table}
\centering
\begin{tabular}{llrr}
\hline
\textbf{Model}&\textbf{Term} & \textbf{\#} & Prop(\%) \\
\hline
-&All Data & 23,162 & - \\
\hline
\multirow{5}{*}{Ours}&Too Long & 233 & 1.0\\
&Power Operator & 51& 0.2 \\ 
\cline{2-4}
&Single& 17,959 & 77.5\\ 
&Multiple& 3,947 & 17.0 \\ 
&Data & 21,906 & \textbf{94.5} \\
\hline
\multirow{2}{*}{WARM}&Data (w/o beam) &-& 14.5 \\
&Data (w/ beam) &-& 80.1 \\
\hline
LBF &-& - & 80.1\\
\hline
\end{tabular}
\caption{Statistics of ComSearch Results.}
\label{tab:search_stat}
\end{table}

\begin{table*}
\centering
\begin{tabular}{crrrr|r}
\hline
\textbf{\#Variable} & Bruce-Force & Removing Brackets & Commutative&ComSearch & Ratio  \\
\hline
1&1&1&1&1&1\\
2&8&8&6&6&1.3\\
3&192&144&108&68&2.8\\
4&9,216&5,184&3,816&1,170&7.9\\
5&737,280&311,040&224,640&27,142&27.2\\
6&88,473,600&27,993,600&19,841,760&793,002&111.6\\
\hline
\end{tabular}
\caption{Empirical Results of Search Space Size.}
\label{tab:compress}
\end{table*}

\section{Analysis on ComSearch}

\subsection{Search Statistics}

We give statistics of ComSearch in Table \ref{tab:search_stat}. Among the 23,162 examples, 233 have more than 6 variables that we filter out, and 51 use the power operation that our method is not applicable. 94.5\% of the examples find at least one equation that can match the answer value, significantly higher than WARM and LBF, which cover only 80.1\% of the examples. 17,959 examples match with only one equation, and 3,947 examples match with two or more equations that need the ranking module to choose the pseudo label further. We show the distribution of these examples in Appendix Section \ref{sec:dis}.

We further break down the recall on different variable sizes in Table \ref{tab:coverage}.
As we can see, when the number of variables grows larger, the recall of LBF drastically collapses, while the recall of our method keeps steady. Sampling based methods cover only a small subset of the equation space and fail to extract candidate equations for larger variable size examples. In contrast, our method can consider a broader range of equation space, which demonstrates the superiority of our enumeration based method.


\begin{table}
\centering
\begin{tabular}{lrrrrrr}
\hline
\textbf{\#Var} &1&2&3&4&5&$\geq$6\\
\hline
LBF&91.5&86.8&88.8&31.1&25.0&38.4\\
Ours&67.0&93.4&96.4&98.1&94.4&73.8  \\
\hline
\end{tabular}
\caption{Result of recall on different variable sizes}
\label{tab:coverage}
\end{table}

\subsection{Eliminating Equivalent Equations in Search Space}

We show the empirical compression of the search space with ComSearch in Table \ref{tab:compress}. As we can see, the compression ratio of ComSearch increases as the variable number grows, up to more than 100 times when the number of variables reaches 6. Previous studies on reducing the redundancy of equivalent expressions consider a limited set of rules, such as removing brackets \cite{roy-roth-2015-solving} and Commutative Law \cite{wang2018translating}. We also show the results of considering removing brackets, where $-$/$\div$ can not be the children node of $+$/$*$, which is the compression considered in \citet{roy-roth-2015-solving}; and Commutative Law, which is the compression considered in \citet{wang2018translating}. Although the two methods can compress the search space to some extent, there is a large gap between their compression efficiency and ours, up to more than 20 times when the number of variables reaches 6.

The size of the Bruce-Force search space could be directly calculated, which is $n!*(n-1)!*4^{n-1}$. If we consider the exponential generating function of $card(S_n)$,  based on Smooth Implicit-function Schema, we can have an approximation of $S_n$: $card(S_n) \sim C * n^{n-1}$, which shows our searching method compresses the search space more than exponential level. We give proof in appendix Section \ref{sec:proof}.

\subsubsection{Advantages of Enumeration without repeat}

The most important core of our approach is that it explicitly points out the \textit{false-matching} problem because it can enumerate a wide range of equations while ensuring each equation holds an independent mathematical reasoning logic. Sampling methods can only sample a small set of equations that may neglect other potential candidates. 

Compared to other enumeration methods, despite the enumeration efficiency, 
Comsearch ensures the enumeration is among non-equivalent equations, so collecting more than one candidate equation for one example shows that there exists more than one mathematical reasoning logic that could reach the annotated answer value. However, only one of the reasoning logic could be true, which elicits the \textit{false-matching} problem. Even if we add more rules to compress the search space, as long as the non-equivalency of different equations cannot be ensured, we cannot differ \textit{false-matching} and multiple expressions of the same mathematical reasoning logic.

\section{Experiments}

\subsection{Dataset and Baselines}

We evaluate our proposed method on the Math23K dataset. It contains 23,161 math word problems annotated with solution expressions and answers. We only use the problems and final answers. We evaluate our method using the train-test split setting of \citet{wang2018translating} by the three-run average. 

We compare our weakly-supervised models' math word problem solving accuracy with two baseline methods. 

\citet{chatterjee2021weakly} proposed \textbf{WARM} that uses RL to train the candidate generation model with the reward of whether the value of the equation is correct. Since the reward signal is sparse due to the enormous search space, the top1 accuracy of the candidate generation model is limited, and they use beam search to search for candidates further.

 \citet{Hong_Li_Ciao_Huang_Zhu_2021} proposed \textbf{LBF}, a learning-by-fix algorithm that searches in neighbour space of the predicted wrong answer by random walk and tries to find a fix equation that holds the correct value as the candidate equation. \textit{memory} saves the candidates of each epoch as training data.

\begin{table}[t]
\centering
\begin{tabular}{p{60pt}|p{60pt}p{60pt}}
\hline
\textbf{Model} & \textbf{Valid(\%)}& \textbf{Test(\%)} \\
\hline
\multicolumn{3}{l}{\small{\textit{GTS based}}}\\
\hline
WARM&-& 12.8\\
\textit{+beam}&-& 54.3 \\
\hline
LBF$\dag$ &57.2($\pm0.5$)& 55.4($\pm0.5$)\\ 
\textit{+memory}$\dag$ &56.6($\pm6.9$)& 55.1($\pm6.2$) \\

\hline
Ours$\dag$&\textbf{61.0($\pm0.3$)}& \textbf{60.0}($\pm0.3$)\\

\hline
\textit{Supervised}$\dag$ &- &75.6\\
\hline\hline
\multicolumn{3}{l}{\small{\textit{G2T based}}}\\
\hline
WARM & - & 13.5 \\
\textit{+beam}&-& 56.0\\
\hline
Ours$\dag$&\textbf{61.7($\pm1.1$)}& \textbf{60.5}($\pm0.6$)\\
\hline
\textit{Supervised}$\dag$&- &77.4\\

\hline
\end{tabular}
\caption{Results on Math23K. $\pm$ denotes the variance of 3 runs for valid/test. \textit{Supervised} denotes full supervision upper bound. $\dag$ denotes the results of our implementation, other results are from the original paper.}
\label{tab:result}
\end{table}

\begin{table}[t]
\centering
\begin{tabular}{lcc}
\hline
\textbf{Model} & \textbf{Valid(\%)} & \textbf{Test(\%)} \\
\hline
Proposed Method & 61.0&60.0\\
w/o Multiple Data &58.9& 57.5 \\
w/o Ranking &57.3& 56.3 \\
w/o Beam Search & 60.1&59.2\\
\hline
\end{tabular}
\caption{Results of Ablation Study for Ranking. `w/o Multiple Data' denotes only using single candidate pseudo data for training. `w/o Ranking' denotes removing the ranking module and randomly sampling an equation for the examples that match with two or more equations. `w/o Beam search' denotes using the basic ranker for ranking. }
\label{tab:ab}
\end{table}

\begin{table}[t]
\centering
\begin{tabular}{lc}
\hline
\textbf{Model} & \textbf{Micro Eq Acc(\%)} \\
\hline

Single & 81.4\\
Multiple & 2.7 \\
All Data & 23.0\\
\textit{Basic Ranker}(Multiple) &45.6 \\
\textit{Beam Ranker}(Multiple) & 47.7 \\
\textit{Beam Ranker}(All Data) & 76.3 \\
\hline
\end{tabular}
\caption{Equation accuracy of different methods. `All Data' denotes considering both the single and multiple data.}
\label{tab:eq_oracle}
\end{table}

\subsection{Main Results and Ablation Study}

We show our experimental results in Table \ref{tab:result}.
We reproduced the results of LBF with their official code and found that LBF+memory lacks robustness. As we can see in the table, the performance of LBF has high variance on both the validation and test set. For a fair comparison, we additionally ran 5-fold cross-validation setting according to \cite{Hong_Li_Ciao_Huang_Zhu_2021} for our model and LBF+memory with the GTS model. The results show that LBF + memory achieves a cross-validation score of 56.3\% with a variance of $\pm6.2$, while our model achieves a cross-validation score of 59.7\% with a variance of $\pm1.0$, which performs similar to the train-test setting.
We observe that its performance highly relies on the initialization of the model. When fewer candidates are extracted at early-stage training, the performance drops drastically since LBF relies on random walks in an enormous search space. 
Our method achieves state-of-the-art performance and outperforms other baselines up to 3.8\% and 2.7\% on train-test and cross-validation settings. Our method is also more robust with minor variance.  

We perform an ablation study with the GTS-based train-test setting in Table \ref{tab:ab}. \textit{Single Equation} denotes using the 17,959 examples that only match with one equation, the model achieves 57.5\% performance, which is slightly lower than using all data and the ranking module, outperforming other baseline models. The result shows that the examples with only one matching could be considered highly reliable and achieve comparable performance with a smaller training data size.  
We observe a performance drop of at least 2.9\% without the ranking module, showing that our ranking module improves the performance. 
We observe a performance gap of 0.9\% between the two rankers, demonstrating the importance of considering candidate equations from the model prediction.

\subsection{Analysis}

We conduct analysis on GTS train-test setting since the model achieves similar performance compared with G2T and the run time is less.

    \begin{figure}[t]
  \centering
  \includegraphics[width=0.45\textwidth]{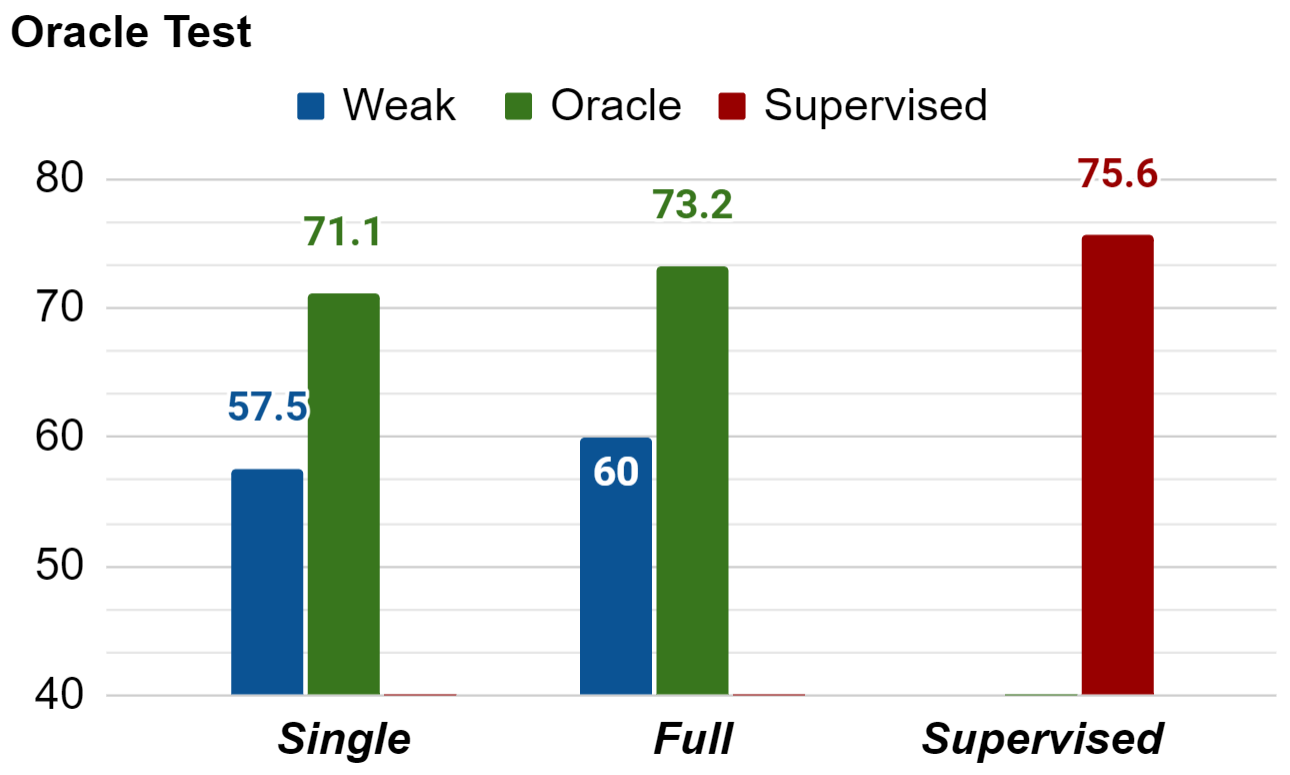}
  \caption{Results of Oracle Test with gold labels.}
  \label{fig:oracle}
  \end{figure}

\begin{table*}[t]
\centering
\begin{tabular}{p{220pt}|p{90pt}p{80pt}p{15pt}}
\hline
\textbf{Text} & \textbf{Candidates} & \textbf{Gold} & \textbf{Ans} \\
\hline

\small{Some children are planting trees along a road every 2 meters. They plant trees on both ends of the road. At last they planted 11 trees. How long is the road?} &\textcolor[rgb]{0,0,0.8}{2*(11-1)}& (11-1)*2 & 20\\
\hline
\small{A library has 30 books. On the first day, $\frac{1}{5}$ of the books were borrowed out. On the second day, 5 books were returned. How many book are there in the library now?} & \textcolor[rgb]{1,0,0}{30 - $\frac{1}{5}$ * 5} & 30*(1-($\frac{1}{5}$)) + 5& 29\\
\hline
\small{Peter and a few people are standing in a line, one person every 2 meters.  Peter found that there are 4 people before him and 5 people after him. How long is this queue?} & \textcolor[rgb]{1,0,0}{4*5-2} , \textcolor[rgb]{0,0,0.8}{(4+5)*2} & 4*2 + 5*2 & 18\\
\hline
\end{tabular}
\caption{Case study of ComSearch. The blue color denotes that the candidate is true-matching and the light red color denotes that the candidate is false-matching.}
\label{tab:case}
\end{table*}
\begin{table*}[t]
\centering
\begin{tabular}{c|rr|rr|rrr}
\hline
&\multicolumn{4}{c|}{Train}&\multicolumn{3}{c}{Test}\\
\hline
  & \multicolumn{2}{c|}{Micro Eq Acc(\%)} &
\multicolumn{2}{c|}{Macro Eq Acc(\%)}& \multicolumn{3}{c}{Ans Acc(\%)}\\
\hline
\textbf{\#Var}&LBF&Ours&LBF&Ours&LBF&Ours& Prop(\%)\\
\hline
1  & 91.8&\textbf{96.3}&\textbf{88.2}&64.9& \textbf{75.0} & 50.0& 1.6\\

2  & 82.9&\textbf{94.8}&78.1&\textbf{88.7}&\textbf{75.2} & 73.4& 33.1 \\
3  & 54.2&\textbf{78.9}&57.4&\textbf{76.1}&56.2 & \textbf{62.9}& 48.5\\
4  &38.0&\textbf{58.0}&13.6&\textbf{57.4}&4.8 & \textbf{25.8}& 12.4\\
5  &8.6&\textbf{31.1}&4.2&\textbf{29.4}&3.2 & \textbf{16.1}& 3.1\\
$\geq$ 6 &5.1&\textbf{50.6}&1.2&\textbf{38.1}&0&\textbf{30.1}&1.3\\
\hline
\end{tabular}
\caption{Results of different variable sizes.}
\label{tab:len}
\end{table*}

\subsubsection{Oracle Test}

While our searching method covers 94.5\% of the training data, as shown in Table \ref{tab:search_stat}, there is still a significant performance gap of more than 15\% between the weakly supervised performance and fully supervised performance, as shown in Table \ref{tab:result}. 
As stated in Section \ref{sec:rank}, we observe that the \textit{false-matching} problem could potentially draw down the performance, which is verified by the effectiveness of the ranking module. 

To further analyze our two modules, we perform two oracle tests for the weakly supervised system. In Figure \ref{fig:oracle}, using the same data examples, we replace the weakly supervised annotations with the supervised gold labels and train the MWP solver. We can observe a performance gap of around 10\% using the same data examples as training data, which indicates that the weakly supervised annotations contain noise. Since all candidate equation annotations have the correct answer, the \textit{false-matching} problem is why this noise exists. The results show that the \textit{false-matching} problem is the critical issue in the weakly supervised setting that causes the performance gap compared to supervised setting.

To investigate the noise in the pseudo training data, we perform an oracle analysis of the \textit{Micro Equation Accuracy} of the pseudo training data. \textit{Micro Equation Accuracy} is defined by what proportion of training instance holds the correct equation solution, which means the instance is not a \textit{false-matching} example. 
In Table \ref{tab:eq_oracle}, we show the results of micro equation accuracy of the training data. 
We check whether the pseudo equation annotations that our system obtains are equivalent to the gold labels for each instance.
We can see that even in the \textbf{Single} data that can only extract one candidate equation, the micro equation accuracy shows there is still noise in the pseudo training data. We show examples in the case study section to explain this problem. The examples that extract more than one candidate have an equation accuracy rate as low as 2.7\%, which makes our ranking system essential. Benefiting from the ranking system, the multiple candidate data can achieve a higher equation accuracy rate. The Beam ranker performs better than the basic ranker considering beam search results.

\subsubsection{Case Study}

We conduct a case study for ComSearch on three examples to further discuss the strengths and limitations of the method in Table \ref{tab:case}. The first example extracts only one candidate equation; although the written expression is different from the gold label, the two equations are equivalent, and the candidate is true-matching. 
The second example extracts only one candidate equation; the \textit{false-matching} candidate coincidentally equals the correct answer with this set of variable numbers. However, the candidate expression and gold label expression are not equivalent. The algorithm reaches a candidate at the stage of using all numbers and does not further search for candidates that use the constant number 1. 
The third example extracts two candidate equations, while only $(4+5)*2$ holds the correct mathematical knowledge. The two candidates appear at the same searching stage, and such \textit{false-matching} cannot be avoided by Comsearch, where we need the ranker to help filter out the \textit{false-matching} noise. In this example, the two rankers both select the correct label.

\subsubsection{Study on Number of Variables}



The distribution of different variable size instances in Math23K dataset is imbalanced, so we further break down the performance of different variable sizes compared with LBF in Table \ref{tab:len}. The \textit{Micro Equation Accuracy} shows our method can extract higher quality pseudo data for all variable sizes compared to previous sampling based methods, especially for examples with more variables.

The recall of candidate extraction methods is another important factor that affects performance. Therefore, in addition to \textit{Micro Equation Accuracy}, we further investigate the \textit{Macro Equation Accuracy} of the two methods, which is defined as equation accuracy on an average of each math word problem. We show that, except for 1 variable, our method has significant advantages over LBF, especially for difficult examples. This demonstrates that our method can effectively extract high equality data of a large quantity. We also show the test answer accuracy of our method and LBF of different variable sizes, which positively correlates with the \textit{Macro Equation Accuracy}. 
Eliminating equivalent equations allows our method to consider the larger search space, while sampling based methods such as LBF limit to a small neighbour space of the model prediction. When the variable number is small, the in-place random walk of LBF can possibly reach the correct equation, that for the examples with 1 or 2 variables, LBF has a slight performance advantage. When the variable number grows larger, as shown in Table \ref{tab:compress}, the gap between the efficiency of our searching method and LBF expands, and our method can consider more equations candidates and achieve higher recall and better recall performance. Moreover, the \textit{false-matching} problem is more severe when there are more variables; ignoring the problem would cause low \textit{Micro Equation Accuracy} and bring in more noise to the pseudo training data.

\section{Related Work}

Early approaches to solving math word problems mainly depend on hand-craft rules and templates \cite{Bobrow:1964:NLI:889266,Charniak:1969:CSC:1624562.1624593}. 
Later studies either rely on semantic parsing~\cite{roy2018mapping,shi2015automatically,zou2019text2math}, or try to obtain an equation template ~\cite{kushman2014learning,roy-roth-2015-solving,koncel2015parsing,roy2017unit}. 
Recent studies focus on using deep learning models to predict the equation template for full supervision setting. 

For weakly supervised setting, \citet{Hong_Li_Ciao_Huang_Zhu_2021} and 
\citet{chatterjee2021weakly} suffers from two major drawbacks. First, they apply equation candidate searching on an enormous searching space, while our method can effectively extract high-quality candidate equations. \citet{Hong_Li_Ciao_Huang_Zhu_2021} results in low robustness and low performance on examples with more variables. \citet{chatterjee2021weakly} results in low coverage of examples that can extract candidate equations. Second, they use all candidate equations for training and neglect the \textit{false-matching} problem, which is the key issue that drags down the model performance in weakly supervised setting, while our ranking module addresses this issue and further boosts the performance.

To eliminate equivalent expressions, \citet{roy-roth-2015-solving} proposed a model that decomposes the equation prediction problem into various classification problems, eliminating some equivalence forms of the equation.
However, the compression is highly integrated with their model and cannot generalize to other models, including the SOTA seq2seq based models. Moreover, it can only cover limited equivalence forms, leaving out various important forms such as Commutative law and Associative law. \cite{wang2018translating} proposed a normalization method for supervised MWP systems that considers Commutative law. The method merges several equivalent expressions into one expression, resulting in the compression of the target equation space. 
However, their method requires bruce-force enumeration before compression, which remains to have high computational complexity. Only limited equivalent forms are considered in both studies, and the equation space is still considerably ample.

Various studies~\cite{kristianto2016mcat,mansouri2021dprl} in ARQMath competition~\cite{mansouri2020finding} and NTCIR benchmark~\cite{zanibbi2016ntcir} have investigated the math retrieval task that retrieves the most related mathematical passage for a question, which have clear semantic meanings given by the textual description. In our ranker setting, the scoring targets, i.e., plane mathematical equations, cannot provide the semantic meanings that contextual embedding similarity based methods used in math retrieval benchmarks require. With fully supervised training data, retrieval-based methods only achieve 40\%~ accuracy~\cite{wang-etal-2017-deep} on Math23K.

Spurious programs in weakly supervised semantic parsing is a close analogy of the false-matching problem, which refers to incorrect programs that lead to correct denotations. The major difference is that the function names of the spurious programs are natural language defined, so the programs have semantic meanings. Extra knowledge bases~\cite{berant-etal-2013-semantic} and lexicon clues~\cite{goldman-etal-2018-weakly} were used to denoise the spurious programs, which is not applicable for complex lexicon patterns MWPs that the solution equation uses operators `$+,-,*,/$' that have no semantic meaning. \citet{pasupat-liang-2016-inferring} uses a small human-annotated dataset for denoising. \citet{guu-etal-2017-language}, which proposes a re-weighted optimization loss for the examples. However, their method relies heavily on hyperparameter tuning and gains negative results on many datasets. Thus these methods are not suitable for the setting in our paper.

\section{Conclusion}
This paper proposes ComSearch, a searching method based on a Combinatorial strategy, to extract candidate equations for Solving Math Word Problems under weak supervision. ComSearch compresses the enormous search space of equations beyond the exponential level, allowing the algorithm to enumerate all possible non-equivalent equations to search for candidate equations. 
We investigate the \textit{false-matching} problem, which is the critical issue that drags down performance, and propose a ranking model to reduce noise. 
Our experiments show that our method obtains high-quality pseudo data for training and achieves state-of-the-art performance under weak supervision settings, outperforming strong baselines, especially for examples with more variables.  

\section*{Limitations}

As we observe from experiments, the performance gap between the most reliable weak data and oracle data is still 10\%, and the noise rate in the pseudo data is still relatively high. This is caused by the stopping strategy of our searching algorithm. Because introducing constant numbers such as 1 and using variables for more than one time would cause meaningless multiple candidate equations (e.g., $n_1/n_1*n_1$, $1*n_1$), we search the equations at various stages: deleting one variable, adding a constant and using one variable multiple times. We stop searching when the stage ends and one equation is obtained. If a more advanced searching strategy that can consider such redundancy could be introduced, the reliability of the weak data could be further boosted.

Meanwhile, our ranking module only denoises multiple candidate equations examples, while the single data also has a volume of noise. We denoise with a simple strategy for one round because we focus on investigating the negative effects of the false-matching problem. For future work, we would consider applying more advanced learning from noise algorithms and denoise more training data.

In Table \ref{tab:coverage}, the results shows a notable discrepancy in the performance of the \#var = 1 when compared to other variable sizes and the baseline. This discrepancy can primarily be attributed to numerous geometrically related questions in the \#var = 1 example set, such as the computation of the volume of a cube $l^3$ given the side length $l$, which is not encompassed by our current search methodology. A straightforward remedy would be to include this equation template in our search when handling \#var = 1; however, we deliberately excluded it from our experiments to maintain consistency across the different variable sizes.

\section*{Acknowledgements}

This work is partially supported by JST SPRING Grant No.JPMJSP2110 and KAKENHI No. 21H00308, Japan.

\nocite{*}

\bibliography{custom}
\bibliographystyle{acl_natbib}
\appendix

\section{Proof for Search Space Approximation}
\label{sec:proof}

Because there is at least one $+$ or $*$ operator for each equation (i.e. $-a-b-c$ is illegal), the target $S_n$ is not symmetric and is hard to directly approximate. We need two assisting targets to form the approximate. This proof majorly relies on \citet{flajolet_sedgewick_2009} and \citet[A140606]{oeis}.

We first consider target $U$ that considers only $+$, $*$ and $\div$ three operators. We sort it into two categories: $U^+$ that the outermost operator is $+$ and $U^{\divideontimes}$ that the outermost operator is $\divideontimes$. Equations such as $\frac{1}{a}*\frac{1}{b-c}$ are still considered illegal. 

$Z$ corresponds to a single variable equation. We can have the construction of $U$:

\begin{align}
    U^+ &= Z + SET_{\geq}(U^{\divideontimes})\\
    U^{\divideontimes} &= Z + (2^2-1)*SET_{=2}(U^+) \\
    &+ (2^3-1)*SET_{=3}(U^+)...
\end{align}

We apply symbolic method to obtain the EGF of the constructions:

\begin{align}
    U^+(z) &= z + \sum_{k\geq2}\frac{1}{k!}[U^\divideontimes(z)]^k\\
    & = z + [e^{U^\divideontimes(z)}-1-U^\divideontimes(z)]\\
    U^\divideontimes(z) &= z + \sum_{k\geq2}\frac{2^k-1}{k!}[U^+(z)]^k\\
    &=z + e^{2U^+(z)} - e^{U^+(z)} - U^+(z)
\end{align}

Meanwhile, we have:

\begin{equation}
    U(z) = U^+(z) + U^\divideontimes(z) - z
\end{equation}

Next, we consider target $T$ that $-a-b-c$ is considered legal. Similarly we define $T^{\pm}$ and $T^\divideontimes$. We consider the construction:

\begin{align}
    T^\pm &= 2Z + SET_{\geq}(T^{\divideontimes})\\
    T^{\divideontimes} &= 2Z + 2[(2^2-1)*SET_{=2}(T^\pm/2)\\
    &+ (2^3-1)*SET_{=3}(T^\pm/2)...]
\end{align}

With symbolic method we have:

\begin{align}
    T^\pm(z) &= 2z + \sum_{k\geq2}\frac{1}{k!}[U^\divideontimes(z)]^k\\
    & = 2z + [e^{T^\divideontimes(z)}-1-T^\divideontimes(z)]\\
    T^\divideontimes(z) &= 2z + 2\sum_{k\geq2}\frac{2^k-1}{k!}[T^\pm(z)/2]^k\\
    &=2z + 2e^{T^\pm(z)} - 2e^{T^\pm(z)/2} - T^\pm(z)
\end{align}

The illegal equations such as $-a-b-c$ in $T$ equals the counts of $a+b+c$, which is actually $U$. So we have:

\begin{equation}
    S(z) = T(z) - U(z)
\end{equation}

We now have the EGF of $S_n$. We can sequentially compute the first few terms of this sequence:

\begin{equation}
    1, 6, 68, 1170, 27142, 793002, 27914126,  ...
\end{equation}

With Smooth implicit-function schema and Stirling approximation function we have, for an EGF $y(z)=\sum_{n\geq0}y_nz^n$, Let $G(z,w)=\sum_{m,n\geq0}g_{m,n}z^mw^n$, thus $y(z)=G(z,y(z))$:

\begin{align}
    n!*[z^n]y(z) &\sim \frac{c*n!}{\sqrt{2\pi n^3}}*r^{-n+1/2}\\
    &\sim \frac{c\sqrt{2\pi nr}}{\sqrt{2\pi n^3}}(\frac{1}{r})^n(\frac{n}{e})^n\\
    &=\frac{c\sqrt{r}}{n}(\frac{n}{re})^n
\end{align}

while r:

\begin{align}
\label{r}
    G(r,s) = s\\
    \frac{\partial G(r,s)}{\partial w}=1
\end{align}

and c:

\begin{equation}
\label{c}
    c = \sqrt{\frac{\partial G(r,s)/\partial z}{\partial^2 G(r,s)/\partial w^2}}
\end{equation}

We still need the two assisting targets to perform the approximation. We have:

\begin{align}
        U^+(z) &= e^{z + e^{2U^+(z)} - e^{U^+(z)} -U^+(z)} \\
        &- e^{2U^+(z)} + e^{U^+(z)} + U^+(z) -1
\end{align}

Let $G(z,w)=z+e^{2w}-e^w-ln(1+e^{2w}-e^w)$, considering \ref{r} and \ref{c}, r, s and c would be constant numbers.

So we have:
\begin{equation}
    n![z^n]U^+(z) \sim \frac{c_1\sqrt{r_1}}{n}(\frac{n}{r_1e})^n
\end{equation}

Similarly we can approximate $U^\divideontimes$, $T^\pm$ and $T^\divideontimes$:

\begin{align}
        n![z^n]U^\divideontimes(z) &\sim \frac{c_2\sqrt{r_1}}{n}(\frac{n}{r_2e})^n\\
        n![z^n]T^\pm(z) &\sim \frac{c_3\sqrt{r_2}}{n}(\frac{n}{r_3e})^n\\
        n![z^n]T^\divideontimes(z) &\sim \frac{c_4\sqrt{r_2}}{n}(\frac{n}{r_4e})^n
\end{align}

So we have:

\begin{align}
    u_n = n![z^n]U(z) \sim \frac{(c_1+c_2)\sqrt{r_1}}{n}(\frac{n}{r_1e})^n\\
    t_n = n![z^n]T(z) \sim \frac{(c_3+c_4)\sqrt{r_2}}{n}(\frac{n}{r_2e})^n
\end{align}

Since $S(z) = T(z) - U(z)$, the subtraction of $u_n$ and $t_n$ would be our approximation. However we observe that $r_1 \gg r_3$, that $u_n$ can be ignored. So we have:

\begin{equation}
        s_n = n![z^n]S(z) \sim \frac{(c_3+c_4)\sqrt{r_2}}{n}(\frac{n}{r_2e})^n
\end{equation}

Q.E.D.
  \begin{figure}[t]
  \centering
  \includegraphics[width=0.4\textwidth]{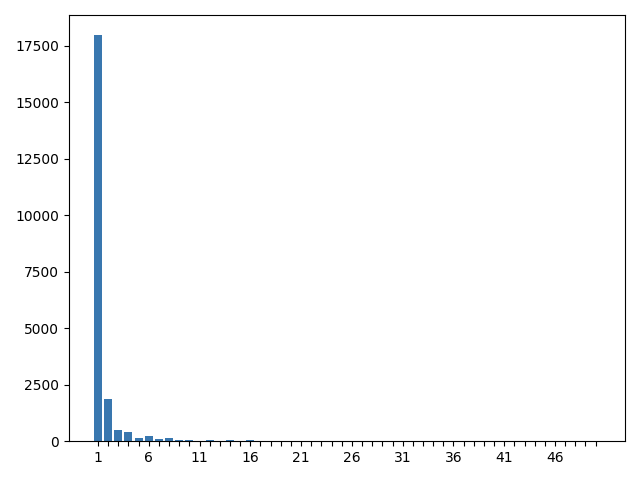}
  \caption{Distribution of Candidate Equation Number.}
  \label{fig:dis1}
  \end{figure}

    \begin{figure}[t]
  \centering
  \includegraphics[width=0.4\textwidth]{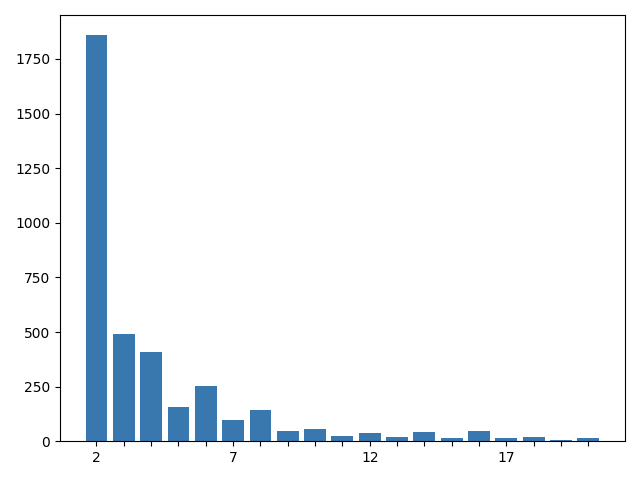}
  \caption{Distribution of Candidate Equation Number.}
  \label{fig:dis2}
  \end{figure}

\section{Distribution of Candidate Equations}
\label{sec:dis}

The largest candidate equation number of one example is 3914. We show the distribution of candidate equations in Figure \ref{fig:dis1} and \ref{fig:dis2}. The x-axis represents the number of candidates, while the y-axis represents the number of examples that have $x$ candidate equations.  We can see from Figure \ref{fig:dis1}, which includes examples that have 1 to 50 candidates, it is a long tail distribution that most examples only have a few candidate equations. From Figure \ref{fig:dis2}, where we zoom in and focus on examples that have 2 to 20 candidates, we can see that there are a lot of examples that have more than 2 candidate equations, and the ranking module is essential.

\section{Experimental Details}

We run our experiments on single card GTX3090Ti, each run takes around 2-3 hours for all models. We did not perform extra hyperparameter searching and use the same hyperparameters as the public release of the two models, except for epoch number which is decided by the validation set. The code is conducted based on Pytorch.

\begin{algorithm}[t]
\caption{\textit{enum\_skel(n)}}\label{alg:cap}
\begin{algorithmic}
\Require $n \geq 1$
\State Initialize empty list skills
  \For{\texttt{$i \leq n$; $i=1$; $i++$}}
  
    \State \textit{left\_list} = \textit{unit\_skel($i$)}
    \State \textit{right\_list} =  \textit{enum\_skels($n-i$)}
    \For {\textit{left} in \textit{left\_list}}
    \For{\textit{right} in \textit{right\_list}}
    \State move the start index of \textit{right} to $i$
    \State \textit{new\_skels} += \textit{left} + \textit{right}
    \EndFor
    \EndFor
    \State \textit{skels} += \textit{new\_skels}
  \EndFor
  \State \textbf{return} \textit{skels}
  \end{algorithmic}
\end{algorithm}

\section{ComSearch Details}
\label{sec:alg}

Considering elements in $S^{\pm}_n$, we can rewrite the equation to $x$. Thus we can form a mapping $g: x \rightarrow  g(x)$ from a general addition equation $x$ to a skeleton structure expression $g(x)$. :

\begin{align*}
    x =& ((x_i\divideontimes(..)) + (x_{j} \divideontimes (..)) + ..) \\&- ((x_{k}\divideontimes(..)) + (x_{l}\divideontimes(..)) + ..)  \\
    g(x)=& (x_i(..))(x_j(..))..\&(x_k(..))(x_l(..))..
\end{align*}
The order of $x_i$ within the same layer of brackets is ignored in $g(x)$, it can deal with the equivalence caused by Commutative law and Associative law. The addition and subtraction terms are split by $\&$, that which can deal with equivalence caused by removing brackets. $g(x)$ is a bijection, so the enumeration problem transforms to finding such skeletons:
\begin{flalign*}
    n=1:&a \\ 
    g^{-1}:&a\\
    n=2:& ab, a\&b, b\&a \\ 
    g^{-1}:& a+b,a-b, b-a\\
    n=3:& abc, a\&(b\&c),(ab)\&c, ...\\
    g^{-1}:& a+b+c,a-(b/c),(a*b)-c, ...\\
    ...\\
\end{flalign*}
The enumeration problem of these structures is an expansion of solving Schroeder's fourth problem~\cite{sch1870}, which calculates the number of labeled series-reduced rooted trees with $n$ leaves. We use a deep-first search algorithm shown in Algorithm \ref{alg:cap} to enumerate these skeletons. It considers the position of the first bracket and then recursively finds all possible skeletons of sub-sequences of the variable sequence $\mathcal{X}=\{x_k\}_{k=1}^i$~\cite{wang2021}.

While considering such skeletons could enumerate all unique expressions, equations have at least one element on the left of $\&$ in our target domain and do not start with $-$ or $\div$. We further extend the algorithm to consider these cases.
To be noticed, because there is at least one $+$ or $*$ operator for each equation, the left side of $\&$ must not be empty while the right part has no restrictions. Thus we define the \textit{unit\_skel($i$)} equation to return possible skeletons with only one or none $\&$ and no brackets. This constraint is equivalent to finding non-empty subsets and their complement of the variable sequence $\mathcal{X}$. 
We can use Algorithm \ref{alg:cap} to perform the enumeration of such skeletons, except for defining two different \textit{unit\_skel($i$)} to support the enumeration of subtraction and division operation. 
The enumeration algorithm of non-empty subsets is trivial and omitted here. 
\begin{equation}
    unit\_skel_{div}(i) = \{(A\&\overline{A})|A\subseteq \mathcal{X};A \neq \emptyset \}
\end{equation}
\begin{equation}
\begin{split}
    unit\_skel_{sub}(i) =&\\
    \{((a(A-a))\&&\overline{A-a})|A\subseteq \mathcal{X};a\in A \}
\end{split}
\end{equation}

We transform the skeletons back to equations to obtain all non-equivalent equations $S_n$. Such enumeration considers absolute values and omits pairs of solutions that are opposite to each other. To search effectively, for the equations that contain subtraction, we add their opposite equation to the searching space.

\end{document}